\documentclass[journal]{IEEEtran}
\usepackage{colortbl}
\usepackage{graphicx}
\ifCLASSINFOpdf
\else
\fi 

\begin{document}

\title{Embodied Edge Intelligence Meets Near Field Communication:  
Concept, Design, and Verification}
\author{Guoliang Li, Xibin Jin, Yujie Wan, Chenxuan Liu, Tong Zhang, Shuai Wang, and Chengzhong Xu
\vspace{-0.1in}
\thanks{
This work is supported by the National Natural Science Foundation of China (Grant No. 62371444), the Open Research Fund of National Mobile Communications Research Laboratory of Southeast University (Grant No. 2025D07), and the Shenzhen Science and Technology Program (Grant No. JCYJ20241202124934046, RCYX20231211090206005).

Guoliang Li and Chengzhong Xu are with the Department of Computer and Information Science, University of Macau, Macau, China.

Xibin Jin is with School of Electronic and Information Engineering, South China University of Technology, Guangzhou, China.

Yujie Wan, Chenxuan Liu, and Shuai Wang are with the Shenzhen Institutes of Advanced Technology, Chinese Academy of Sciences, Shenzhen, China.
Yujie Wan is also with the Southern University of Science and Technology.
Chenxuan Liu is also with the University of Chinese Academy of Sciences. 

Tong Zhang is with Guangdong Provincial Key Laboratory of Aerospace Communication and Networking Technology, Harbin Institute of Technology, Shenzhen, China, and also with National Mobile Communications Research Laboratory, Southeast University, Nanjing, China.

Guoliang Li and Xibin Jin contributed equally.

Corresponding author: Tong Zhang (tongzhang@hit.edu.cn) and Shuai Wang (s.wang@siat.ac.cn).}
}
 
\maketitle
 
\begin{abstract}
Realizing embodied artificial intelligence is challenging due to the huge computation demands of large models (LMs). 
To support LMs while ensuring real-time inference, embodied edge intelligence (EEI) is a promising paradigm, which leverages an LM edge to provide computing powers in close proximity to embodied robots. 
Due to embodied data exchange, EEI requires higher spectral efficiency, enhanced communication security, and reduced inter-user interference. 
To meet these requirements, near-field communication (NFC), which leverages extremely large antenna arrays as its hardware foundation, is an ideal solution. 
Therefore, this paper advocates the integration of EEI and NFC, resulting in a near-field EEI (NEEI) paradigm. 
However, NEEI also introduces new challenges that cannot be adequately addressed by isolated EEI or NFC designs, creating research opportunities for joint optimization of both functionalities.
To this end, we propose radio-friendly embodied planning for EEI-assisted NFC scenarios and view-guided beam-focusing for NFC-assisted EEI scenarios.
We also elaborate how to realize resource-efficient NEEI through opportunistic collaborative navigation.
Experimental results are provided to confirm the superiority of the proposed techniques compared with various benchmarks.
\end{abstract}

 
\IEEEpeerreviewmaketitle

\section{Introduction}

\begin{figure*}[!t]
    \centering
    \includegraphics[width=1.0\textwidth]{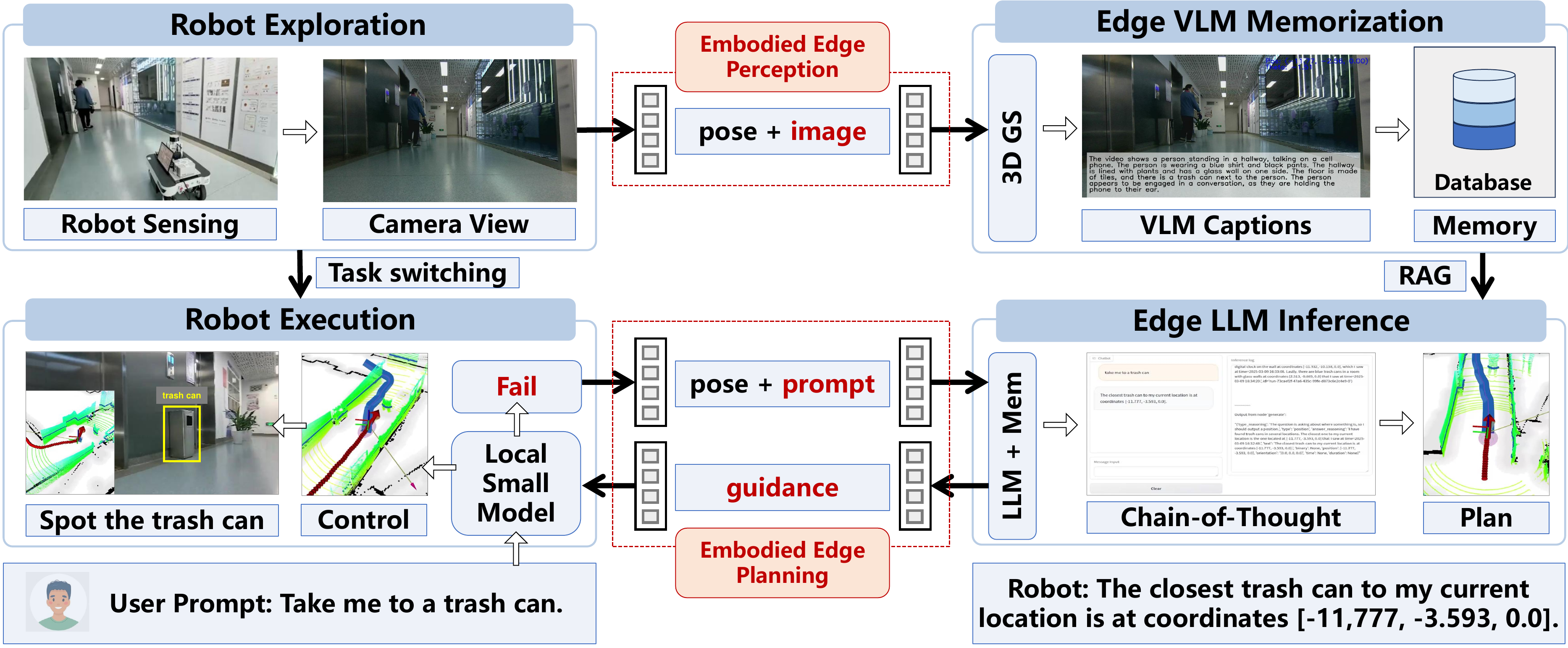}
    \caption{Embodied edge intelligence for object search, which consists of embodied edge perception and embodied edge planning. Both rely on collaboration between an embodied robot with multi-modal sensors and an edge server with LLMs and VLMs.}
    \label{fig1}
\end{figure*}

Enabling an autonomous robot to interpret human prompts, understand tasks, make plans, and take actions is a great challenge due to the huge semantic and geometric search space \cite{nygaard2021real}. 
Recently, however, this situation has changed due to the breakthrough of large models (LMs) \cite{zhao2023large}, which can effectively reduce the search space by leveraging the commonsense reasoning capabilities \cite{anwar2024remembr}.
The convergence of robotics and LMs gives rise to a new paradigm termed embodied artificial intelligence (AI).

Due to the huge computation demands, LMs often reside on the cloud \cite{qu2025mobile}.
The unstable communication between robots and the cloud contradicts to the real-time requirement of embodied AI \cite{qu2025mobile}.
To address this problem, embodied edge intelligence (EEI) emerges, which adopts an \emph{LM edge} to provide computing powers in close proximity to \emph{embodied robots}. 
However, EEI requires robot-edge collaboration \cite{li2024edge}, and the complex embodied perception and planning tasks make EEI more ``communication-hungry'' and ``latency-sensitive'' than traditional edge intelligence \cite{wang2023federated}, calling for more advanced wireless technologies. 

To meet the above requirements, this paper advocates the adoption of near-field communication (NFC) to support  EEI, resulting in a near-filed EEI (NEEI) paradigm.
Particularly, NFC adopts an increased number of antennas (e.g., $\geq 500$) and carrier frequency (e.g., over $10$\,GHz), such that its electromagnetic waves propagate as spherical waves \cite{cui2022near,liu2023near,wang2025new} instead of planar waves in conventional far-field communication (FFC) 
\cite{mohammadi2024ten}.
Built upon these new characteristics, NFC not only offers angle-domain but also distance-domain resolution, thereby providing higher spectral efficiency, enhanced communication security, and reduced inter-user interference \cite{zhuo2024extremely,chen2025hybrid}.

However, NEEI involves interdependency between EEI tasks and NFC conditions. 
Existing studies on NFC \cite{cui2022near,liu2023near,zhuo2024extremely,chen2025hybrid} or EEI \cite{nygaard2021real,zhao2023large,anwar2024remembr,qu2025mobile} ignore these features, resulting in isolated designs and degraded end-to-end performance. 
To fill this gap, this article integrates EEI features and NFC propagation characteristics for \emph{intertwined designs and optimization}.
Frameworks of EEI-assisted NFC and NFC-assisted EEI are presented. 
Based on these frameworks, methods of radio-friendly embodied planning (REP) and view-guided beam-focusing (VBF) are proposed. 
The REP scheme plans an active path for the robot to enjoy better NFC beam-focusing at new positions. 
The VBF scheme distinguishes the heterogeneous view contributions of different frames to obtain the best scene reconstruction quality under stringent NFC resource constraints.
In the cases when resource-efficient NEEI is required, we further illustrate how to reduce NFC overheads by avoiding non-beneficial collaboration via opportunistic collaborative navigation (OCN).
Lastly, we adopt high-fidelity simulations and real-world datasets to verify the above principles and methods. Experiments demonstrate the effectiveness of our NEEI, and confirm the necessity of mutual assistance between EEI and NFC.

\section{Embodied Edge Intelligence:\\Concept and Requirements}

\subsection{Embodied Edge Intelligence}

Embodied AI is an integrated robotics and AI system with multi-modal sensors, onboard computers, actuators, and communications units \cite{nygaard2021real}. 
In contrast to conventional robotic systems that require humans to provide explicit goals (e.g., navigation goals, manipulation trajectories), embodied AI allows the robots to interpret implicit human prompts (e.g., languages, pictures), understand tasks, make plans, and generate goals by themselves \cite{anwar2024remembr}.
This significantly enhances the smoothness of human-machine interactions.
For instance, in shopping malls, users typically cannot specify destination coordinates, but can describe their requirements, such as ``\texttt{take me to a Chinese restaurant}''.

Embodied AI faces a vast search space, making their practical deployment highly challenging. 
Recently, LMs, which have demonstrated strong reasoning capabilities, are adopted for reducing the search space.
These LMs can be large language models (LLMs), e.g., GPT, Llama, DeepSeek, or vision language models (VLMs), e.g., CLIP, Gemini, LLaVA. 
This makes ``LMs + Robotics'' a core paradigm for embodied AI \cite{zhao2023large}. 

However, different from small models (SMs), LMs make onboard computing resources insufficient for real-time inference. 
Cloud LM deployment may result in unstable connections, which contradicts with the low-latency communication requirements from embodied AI \cite{li2024edge}.  
To the end, embodied edge intelligence (EEI), which integrates lightweight robot SMs with powerful edge LMs, can be utilized to ease the conflict between real-time requirements and resource-hungry LM services.

\begin{figure*}[!t]
    \centering
    \includegraphics[width=1.0\textwidth]{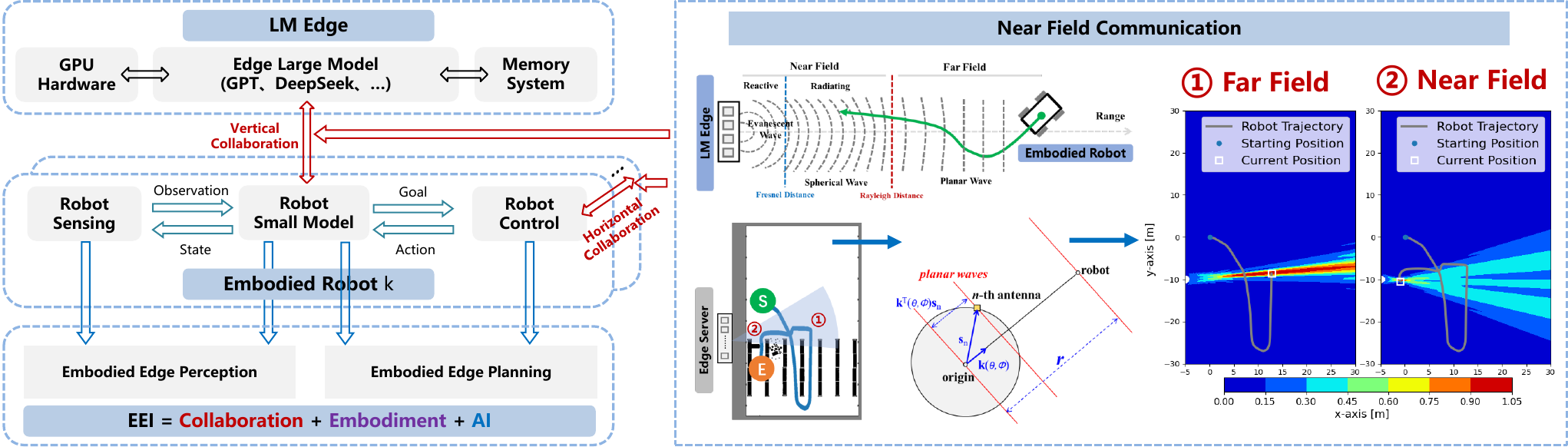}
    \caption{System architecture of NEEI and the spherical-wave propagation characteristics of NFC.}
    \label{fig2}
\end{figure*}

\subsection{Use Case Study}

To illustrate how EEI works, we conduct a prototype experiment based on ROS2 in an SIAT office environment as shown in Fig.~\ref{fig1}.
The embodied robot is an Agilex Ranger autonomous vehicle, which has a 3D lidar, an RGBD camera, and an onboard Intel i5 computer. The LM edge is a Linux workstation with a 3.7 GHZ AMD CPU and an NVIDIA A6000 GPU. The robot and edge are connected over wireless channels. 

In this experiment, the user wants to dispose of garbage, and prompts the robot: ``\texttt{Take me to a trash can}''. 
The robot feeds this prompt, together with its camera images, into the local \texttt{CLIP} model, to check if any related object is within the current or historical views. 
If so, the position of the grounded object in the camera plane is projected to the three-dimensional (3D) space, which is then passed to the back-end model predictive controller for generating actions \cite{li2024edge}. 
The task is deemed successful if the robot and the target object are with a predefined distance threshold (e.g., $5$ meters). 

If the local \texttt{CLIP} fails in returning a valid answer, a robot-edge collaboration event will be triggered. 
Specifically, the robot first uploads the abstract information (including robot pose and user prompt) to the edge server, which feeds the abstract into the \texttt{Codestral-22B} model for LLM agentic inference.
Since the LLM agent does not possess spatial cognition, it needs to query the spatial memory stored in the vector database \cite{anwar2024remembr}, so as to generate goals in the 3D space. 
After chain-of-thought (CoT) inference, the memory-assisted LLM identifies two trash cans, and selects the nearest one located near the elevator. 
The final output is given by: ``
\texttt{The closest trash can to my current location is at coordinates [-11,777, -3.593, 0.0]}''.
With the current position uploaded by robot and the goal position generated by LLM, 
the trajectory planner (e.g, hybrid A$^*$) computes a shortest path (i.e., a list of waypoints) connecting the two points. 
The path serves as a guidance downloaded by the robot for generating control signals.
This completes the entire \emph{embodied edge planning} procedure.  

On the other hand, the construction of spatial memory at the edge also requires robot-edge collaboration. 
This corresponds to the \emph{embodied edge perception} procedure, which consists of robot exploration and edge memorization steps, as shown in the upper part of Fig.~\ref{fig1}.
First, during the robot exploration step, 
the robot executes simultaneous localization and mapping (SLAM) to obtain a (pose, image) pair at each frame, and explores the operational region based on coverage maximization to obtain a sequence of (pose, image) data. 
Second, these data sequences are uploaded to the edge for VLM memorization, which builds 3D Gaussian splatting (GS) models \cite{kerbl20233d}) that can render images for a sequence of poses. 
The edge then captions the rendered images using \texttt{VILA-3B} trained with Multimodal C4 \cite{anwar2024remembr}, and stores the (pose, description) pairs into the memory database.  
With spatial memory, the LLM agent can map the user prompt to the target pose in 3D space, by first selecting the scene description with the highest similarity, and then associating the pose capturing the scene.

\subsection{Requirements}

Section II-B shows that EEI can be categorized into embodied edge perception and embodied edge planning.
Their challenges are summarized below.
\begin{itemize}
\item
\textbf{Embodied Edge Perception}. 
It involves transmission of multi-modal, heterogeneous, high frequency, high resolution robot sensor data. Establishing low-cost communication between robot and edge is non-trivial due to the \textbf{vast data volume} of these sensor data. 
For instance, during the VLM memorization stage in Fig.~\ref{fig1}, a single 1080P $50$FPS camera may require a data-rate of over $400$\,Mbps.
These data also contains \textbf{human-related private} information, which requires strong data privacy protection \cite{wang2023federated}. 
\item
\textbf{Embodied Edge Planning}. Robots need to send query and receive guidance from the edge with \textbf{ultra-low latency}, so as to make precise and timely actions \cite{li2024edge}. For instance, during the LLM inference stage in Fig.~\ref{fig1}, the exchange of prompt and guidance should be accomplished within tens of milliseconds; otherwise the quality of experience for human-robot interaction will be severely degraded.
However, current communication still faces difficulties in meeting these real-time requirements due to inevitable \textbf{inter-robot interference}.
\end{itemize}
In short, EEI raises three key requirements, i.e, improved spectral efficiency, enhanced communication security, and reduced inter-user interference. 
These requirements can be \emph{simultaneously} satisfied by NFC, giving rise to the NEEI paradigm detailed in the subsequent section.

\begin{figure*}[!t]
    \centering
    \includegraphics[width=1.0\textwidth]{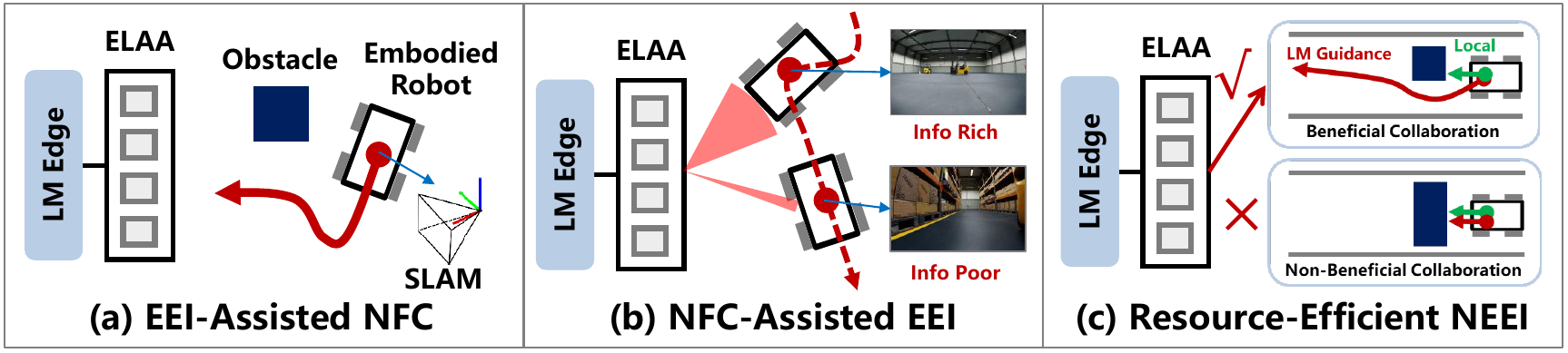}
    \caption{Illustration of the three research opportunities and their differences compared with existing NFC systems.
    }
    \label{fig3}
\end{figure*}

\section{Near-Field Embodied Edge Intelligence}

\subsection{System Overview}

The system architecture of NEEI is shown in Fig.~\ref{fig2}. 
The key is to capitalize on extremely large antenna array (ELAA) as the hardware foundation and leverage NFC to support collaboration between LM edge and embodied robots.
As shown at the right hand side of Fig.~\ref{fig2}, NFC refers to wireless communication operating within the radiative near-field region, characterized by spherical wave propagation of electromagnetic fields, which is in contrast to the planar wave propagation in conventional FFC \cite{wang2025new}. 
In NFC, the linear distance $\|\mathbf{r}-\mathbf{k}^T\mathbf{s}_n\|_2$ under the planar wave assumption would lead to non-negligible error compared to the actual nonlinear propagation distance $\|\mathbf{r}-\mathbf{s}_n\|_2$ of the link between each transmit antenna element and the robot.
In summary, NFC exhibits unique features in three key aspects: 
\begin{itemize}
\item Its electromagnetic waves propagate as spherical waves; 
\item It enables beam focusing in both angular and distance dimensions, as opposed to far-field's angular-only beam steering; 
\item It adopts a significantly increased number of antennas and carrier frequency.
\end{itemize}

These characteristics grant NFC new advantages: 
\begin{itemize}
\item Enhanced spatial resolution through beam focusing, improving spectral efficiency and system throughput; 
\item Reduced information leakage due to concentrated energy distribution, strengthening physical-layer security; 
\item Superior multi-user interference management, allowing multiplexing at identical angles via distance-domain differentiation.
\end{itemize}
It can be seen that the advantages provided by NFC perfectly match the requirements from EEI.

\subsection{NFC versus NEEI: What is the Difference?}

\textbf{Difference 1: Passive NFC versus Active NFC.} 
In NFC systems, users are non-autonomous and passive.
In contrast, users in NEEI are embodied robots that can perceive the environments and make adjustments actively. 
The robot position is controllable, representing an additional system variable to be optimized.

\textbf{Difference 2: Communication oriented versus embodied AI oriented.}
Existing NFC systems \cite{cui2022near,liu2023near,zhuo2024extremely,chen2025hybrid} aim to optimize communication performance (e.g., data-rate, energy efficiency, power consumption). 
However, these metrics are not tailored for domain-specific embodied scenarios. 
In contrast, NEEI aims to explicitly maximize the end-to-end embodied task performance under NFC conditions. 

\textbf{Difference 3: Internet data versus embodied Data.} 
In contrast to traditional NFC that handles Internet data, NEEI needs to support low-latency exchange of embodied data, which is of high frequency, high resolution, and multi-modality (e.g., images, point-clouds, maps).

\begin{figure*}[!t]
    \centering
    \includegraphics[width=0.98\textwidth]{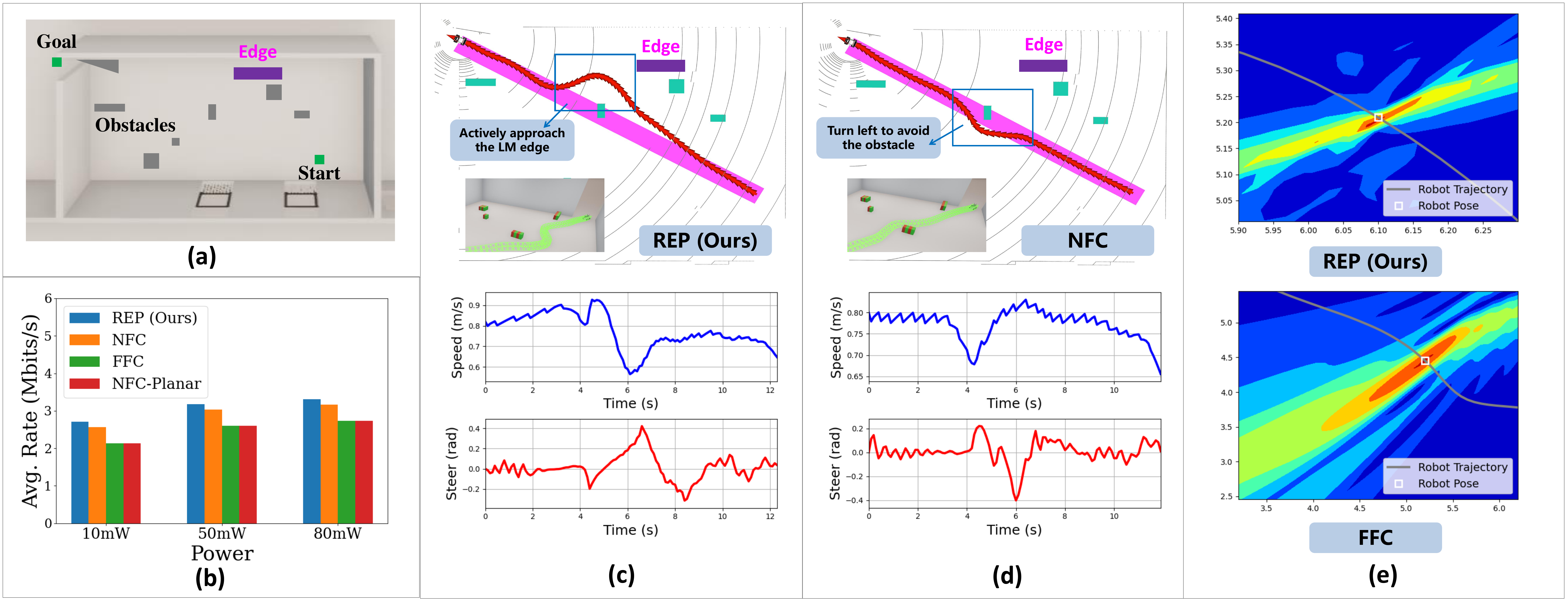}
    \caption{Verification of REP in the EEI-Assisted NFC scenario. (a) Bird's-eye view of the scenario; (b) Comparison of NFC performance; (c) Robot trajectory and control profile of REP; (d) Robot trajectory and control profile of NFC; (e) Visualization of NFC and FFC beamforming. 
    }
    \label{fig4}
\end{figure*}

\subsection{Opportunities and Challenges}

The above differences open up the following opportunities.

\textbf{Opportunity 1}: EEI-assisted NFC, where robots plan radio-friendly trajectories to enhance NFC performance, as illustrated in Fig. 3a.

\textbf{Opportunity 2}: NFC-assisted EEI, where NFC enables robots to perform collaborative embodied tasks, as illustrated in Fig. 3b.

\textbf{Opportunity 3}: Resource-efficient NEEI, where edge-robot collaboration is optimized to minimize communication overhead, as illustrated in Fig. 3c.

However, new features of NEEI lead to new technical challenges, which cannot be tackled by existing methods.

\textbf{Challenge 1}: 
How can the robot plan a path to enjoy better beam spots at positions with reduced propagation loss and interference?

\textbf{Challenge 2}: How can we co-optimize the NFC beamforming and resource allocation across different times, robots, tasks, and modalities to maximize the end-to-end EEI task performance under stringent NFC constraints? 

\textbf{Challenge 3}: How can multi-functional coordination (joint optimization of sensing, communication, computation, and control) reduce the data demands of NEEI?

To tackle these challenges, Sections IV--VI will present potential solutions.

\section{EEI-Assisted NFC}
We begin with EEI-assisted NFC. 
In this scenario, the principle is to leverage robotics capabilities for enhancing the NFC performance metrics. 
\subsection{Problem Description} 
The problem is to plan a radio-friendly path for the robot to enjoy better NFC beam-focusing at new positions.
However, NFC beamforming and robot planning in EEI-assisted NFC are interdependent, for which existing NFC approaches \cite{cui2022near,liu2023near,zhuo2024extremely} become inefficient, as they treat the position of user as uncontrollable. 
To fill the gap, it is necessary to integrate NFC and EEI features for joint optimization. 
Note that different from UAV communication \cite{guo2021uav,chen2025hybrid} under low uncertainties, EEI operates in dynamic scenarios with high uncertainties.
Existing approaches \cite{guo2021uav,chen2025hybrid} calculating an offline trajectory may result in discrepancies between the intended path and robot's actual environments, primarily due to the need to avoid collisions with dynamic obstacles. Existing works \cite{guo2021uav,chen2025hybrid} also overlook the NFC propagation features.

\subsection{Radio-Friendly Embodied Planning}
We propose the REP method that accounts for the real-time dynamic collision avoidance and NFC features. 

First, we model the NFC channel as a function of each robot pose. 
The NFC channel consists of deterministic line of sight (LoS) and stochastic NLoS elements \cite{liu2023near}.
For the LoS part, given a certain antenna configuration (e.g., antenna aperture), we adopt physical model to map the robot location to a corresponding channel coefficient. For the NLoS part, they rely on statistical models with parameters (e.g., mean, variance) learnt from historical measurements or pilots.

Second, given a sequence of robot poses, we incorporate the above NFC-pose channel function into a Markov decision process (MDP). 
This MDP describes the relation between NFC channels and robot actions under certain robot dynamics (e.g., differential-drive, Ackermann) \cite{han2023rda}. 
Based on MDP, we can jointly optimize the cross-layer variables of NFC beam-focusing vectors and robot actions. 

Third, the robot must strictly avoid collisions with all obstacles within the occupancy map. 
The collision avoidance constraint is formulated by computing the minimum distance between two polyhedrons, and can be tackled by parallel optimization algorithms. 
When the map is downloaded from the edge server, the constraint additionally becomes a function of the downlink NFC performance, introducing a latency-dependent safety constraint.

\subsection{Experiments}
We evaluate the proposed REP in Carla simulation, which provides high-fidelity rendering qualities, driving behaviors, and software interfaces \cite{wang2023federated}. 
As shown in Fig.~4a, we consider an indoor scenario with $6$ obstacles, and the robot needs to move outside the room. 
For communication, the LM edge is marked as purple rectangle and equipped with the uniform linear array (ULA) as in \cite{liu2023near}. 
The communication bandwidth is $200$~KHz and noise power is $-80$\,dBm. 
Pathloss is computed based on the positions of edge and robot, with pathloss exponent equal to $2$. 
For robot planning, the sample time is $0.1$\,s, the length of planning horizon is $20$, the safety distance is $0.1$\,m, and the reference speed is $0.5$\,m/s.

We compare our REP with the following baselines ($N_t$ and $f_c$ denote the number of antennas and carrier frequency): 
1) NFC \cite{cui2022near}, which adopts $N_t=640$ and $f_c=30$\,GHz;
2) FFC \cite{mohammadi2024ten}, which adopts $N_t=32$ and $f_c=1.5$\,GHz;
3) NFC-Planar, which is a variant of NFC but adopts planar-wave channel model for beamforming design. 
These baselines as well as the antenna configurations are also adopted in Sections V and VI.

As illustrated in Fig.~4b, the proposed REP demonstrates superior performance compared to the NFC scheme in terms of the average data collection rate. 
This is attributed to the cross-layer optimization of NFC beam-focusing and robot active planning, and the associated gain is defined as \textbf{EEI-assisted NFC gain}. 
In addition, the performance of NFC is better than FFC, since NFC enables beam focusing in both angular and distance dimensions, as opposed to angular-only beam steering of FFC, which corroborates our discussions in Section III-A. 
The NFC-Planner scheme leads to the worst performance, which demonstrates the necessity of applying nonlinear channel model in NEEI systems.

To obtain deeper insights, it is observed from Fig.~4c and Fig.~4d that, under the proposed REP scheme, the robot slows down and turns right at $t=5s$ to actively approach the LM edge for better NFC quality, thereby enhancing the NFC data harvesting, which corroborates Fig.~4b. 
In contrast, the NFC plans a left-turn trajectory to avoid obstacles as shown in Fig.~4d and ignores the potential EEI assistance for NFC.  
We also provide the beamforming visualization in Fig.~4e. 
It can be seen that empowered by NFC, our REP method focuses almost all energies into a spot, as opposed to the beam steering in the FFC case.

\begin{figure*}[!t]
    \centering
    \includegraphics[width=0.98\textwidth]{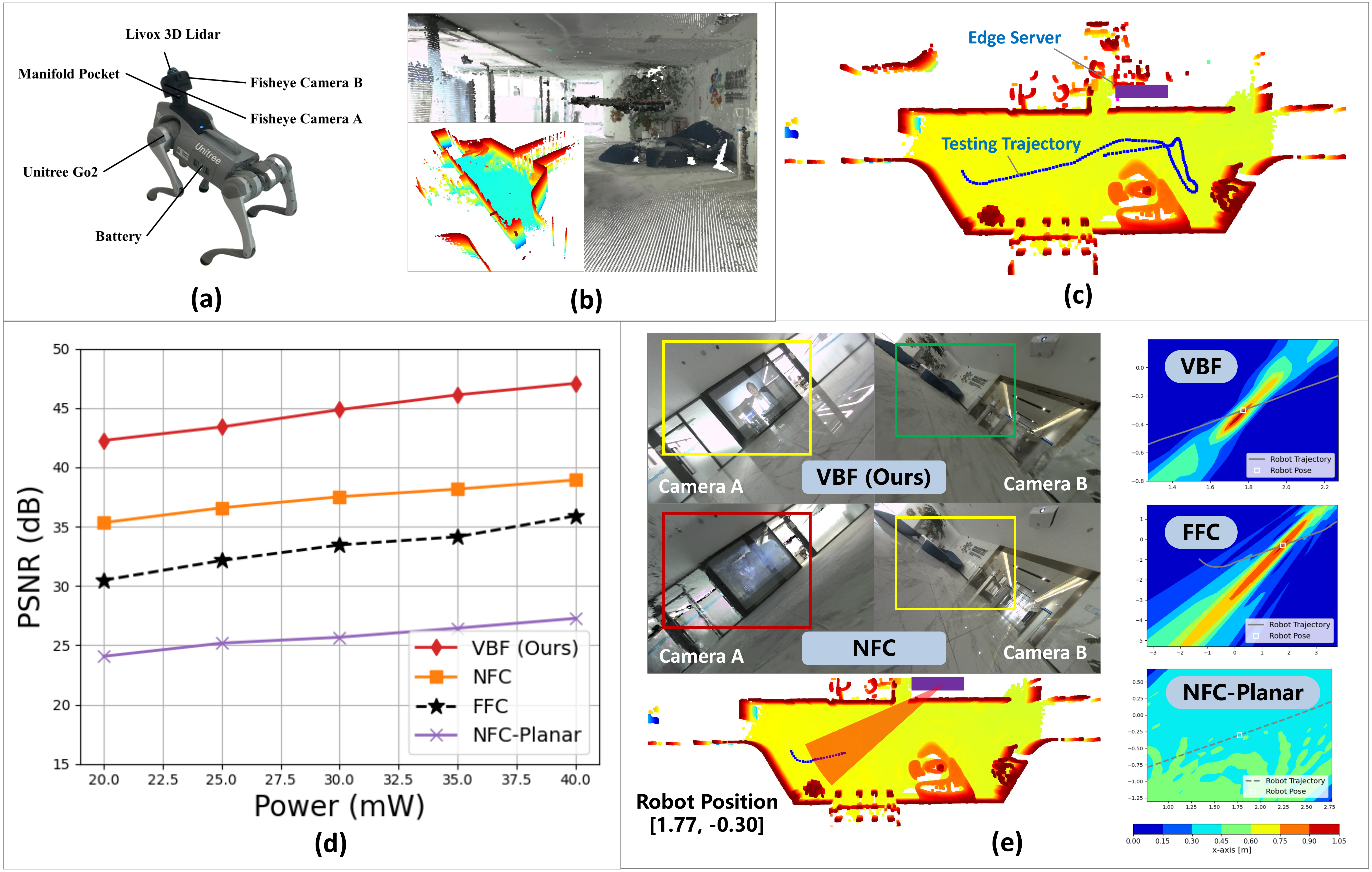}
    \caption{Verification of VBF in NFC-assisted EEI scenarios. (a) Robotic dog platform; (b) The pretrained GS model; (c) The robot exploration trajectory; (d) Quantitative comparison between the proposed VBF and other benchmarks; (e) Visualization of rendered images and beamforming gains. 
    }
    \label{fig5}
\end{figure*}

\section{NFC-Assisted EEI}

To elucidate the underlying principle of NFC-assisted EEI, we focus specifically on 3D GS reconstruction as a representative use case. This implementation demonstrates how NFC facilitates real-time image acquisition from mobile robotic platforms for assisting 3D vision LMs at the edge.

\subsection{Problem Descripion}
While GS can render a photo-realistic view from the robot’s pose, it may involve discrepancies compared to the actual environments (i.e., termed “memory bias”), primarily due to the variations in illuminations and changes of environments \cite{kerbl20233d}.
Therefore, edge GS allows robot images to be opportunistically uploaded to complement the GS model. 
However, this problem aims to maximize the rendering qualities of edge GS instead of the NFC performance. This implies that image frames should be prioritized according to their view contributions to GS rendering. 
Intuitively, more communication resources should be allocated to those unexplored scenes that have higher entropy. However, current NFC resource allocation methods \cite{cui2022near,liu2023near,zhuo2024extremely,chen2025hybrid} focus on the objective of communication throughput and treats data equally.
The previous general-purpose learning objective in edge intelligence \cite{wang2023federated} cannot reflect the 3D geometric characteristics and fail to account for the non-uniform contribution of images to the rendering quality.
Therefore, how to define a proper objective function for edge GS still remains an open question.

\subsection{View-Guided Beam Focusing}
We propose the VBF method, which unifies view-point contribution, NFC beam focusing, and cross-frame power allocation into a single optimization framework, to obtain the best scene reconstruction quality under stringent NFC resource constraints. 

First, a novel GS-oriented objective function is formulated to distinguish the heterogeneous view contributions of different frames. This objective function,
which is built upon the uncertainty sampling theory and the vanilla GS loss
function, enables the edge server to maximize the information gain brought by the collected sensor data. 
Second, the new objective function is integrated with the NFC conditions for effective cross-layer optimization. 
This leads to a joint frame selection and power allocation problem that is solved by mixed integer nonlinear programming. 
Third, the GS-oriented objective, being a function of images, is inaccessible prior to data transmissions. To this end, we can adopt a sample-then-transmit strategy, which first samples a small subset of representative data and then transmits the subset as pilot images for validation and loss estimation.
We can also prebuild a GS-loss map to mark the low-quality rendering regions. 

\begin{figure*}[!t]
    \centering
    \includegraphics[width=0.98\textwidth]{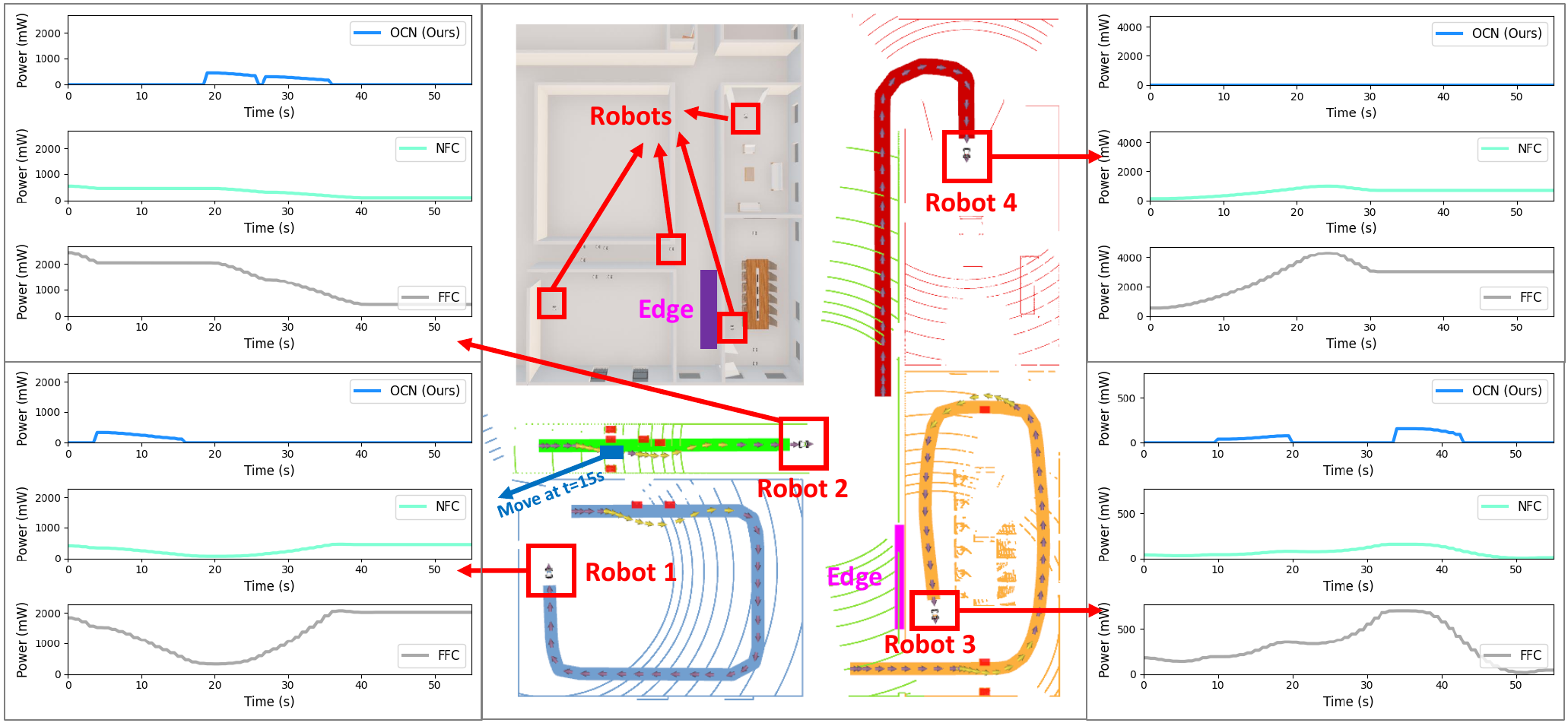}
    \caption{Verification of OCN in a multi-robot indoor scenario in Carla-ROS simulation.}
    \label{fig6}
\end{figure*}

\subsection{Experiments}
We conduct experiments using the Unitree Go2 robotic dog platform shown in Fig.~\ref{fig5}a in an office environment. 
We collect $400$ images for training a GS model and the trained GS model is shown in Fig.~\ref{fig5}b.
We then navigate the robot along the trajectory in Fig.~\ref{fig5}c, capturing $370$ images.
The task is to decide which images should be uploaded to LM edge via uplink NFC.
The LM edge is located at $[5,4]$ and marked as purple rectangle.
The bandwidth is $10$\,MHz, noise power is $-80$\,dBm, pathloss exponent is $3$, and each image is $1.73$\,MBytes.

We compare the peak signal-to-noise ratios (PSNRs) (i.e., key metric of GS) of different schemes in Fig.~\ref{fig5}d. 
It can be seen that no matter how robot power changes, the proposed VBF outperforms existing NFC by at least $6$\,dB. 
This is because our VBF would prioritize NFC powers towards more valuable data frames that involve discrepancies between the captured image and the GS rendered view. This is the case of camera A in Fig.~\ref{fig5}e. 
Otherwise, if the edge GS already renders a high-fidelity picture (e.g., marked in green box in camera B of Fig.~\ref{fig5}e), our VBF would not trigger its upload. 
In contrast, the NFC scheme uploads the image B for NFC throughput maximization, resulting in blurred defects in camera A, where the dynamic screen is severely damaged as marked by red box.
This demonstrates the effectiveness of the \textbf{embodied-oriented principle}. 
To understand why the proposed VBF outperforms FFC and NFC-Planner, we also illustrate their beam-forming heat maps at robot position $[1.77,-0.30]$ in Fig.~\ref{fig5}e, where the beam direction is to the south-west.
Empowered by NFC, our method focuses almost all energies into a spot.
In contrast, the FFC method forms a beam steer towards the target angle, with a large power leakage along the line. 
Finally, the NFC-Planner scheme results in an improper beam-focusing position.

\section{Resource-Efficient NEEI}

While Sections IV and V have demonstrated how EEI and NFC can mutually enhance system performance, practical implementation requires additional consideration of resource efficiency to prolong robotic system lifetimes. To achieve this, a holistic optimization approach should be adopted across complementary functionalities including sensory processing, computation offloading, and control strategy optimization.

\subsection{Problem Description}
Navigating robots to the desired goal while ensuring safety is a foundation task \cite{han2023rda}. 
Traditional navigation models lead to inaccurate trajectories in cluttered environments.
Emerging LM-guided navigation \cite{anwar2024remembr} has demonstrated remarkable intelligence and accuracy, but is hard to be deployed at the robot due to the limited onboard computing power. 
Indeed, insufficient planning frequency may result in delayed actions and potential collisions.
To this end, a promising solution is to leverage collaborative planning by offloading computationally intensive planning to the proximal LM edge. 
However, the key challenge lies in determining the optimal timing for engaging the edge planning. 
This is crucial since an improper collaboration timing may lead to a waste of NFC and edge computing resources, without yielding substantial navigation improvements.

\subsection{Opportunistic Collaborative Navigation}

We propose an OCN method to avoid non-beneficial collaboration, enabling adaptive computation splitting between robots and servers.
The OCN deploys a local planning model (e.g., path following) with minimal computational requirements at the robot. 
Meanwhile, an advanced planning model (e.g., LM-guided collision avoidance) is deployed at the edge, facilitating aggressive behaviors that can navigate the robot through clutter (e.g., narrow gaps). 

We define the expected collaboration gain as the difference between trajectories planned by edge and robot.
Furthermore, model predictive collaboration (MPCO) is employed to maximize the expected collaboration gain while satisfying computation and communication resource constraints. 
In contrast to model predictive control \cite{han2023rda} that merely considers robot states and actions, MPCO incorporates collaboration status into decision making to identify robots that would most benefit from collaboration. 
This results in a bilevel mixed-integer nonlinear programming algorithm, 
where the upper-level problem orchestrates the computation and communication resources, and the lower-level problem solves a set of collision avoidance constraints.

\subsection{Experiments}
We implement OCN in a multi-robot indoor scenario illustrated in Fig.~\ref{fig6}, where robot 1 navigates in an office room with target path marked in blue, robot 2 navigates in a corridor with target path marked in green, robot 3 navigates in a conference room with target path marked in orange, and robot 4 navigates from a corridor to a lounge with target path marked in red. 
The edge server is located at $[0,0]$ and marked as a purple rectangle, while the obstacles are marked as red boxes. 
The noise power is $-100$\,dBm and pathloss exponent is $2$.
To collaborate with the server, the uplink SINR of the associated robot should be above $20$\,dB.
The hyperparameters for motion planning are the same as those in Section IV.

The trajectories of $4$ robots are shown in Fig.~\ref{fig6}, where the purple trajectories are generated by robot and the yellow trajectories are generated by edge. It can be seen that robot 4 encounters no obstacles along its target path; hence it keeps on moving using the local model until reaching the goal at $t=30\,$s. 
Therefore, our proposed OCN disconnects this robot during the entire task execution process, leading to zero power consumption as shown at the upper right of Fig.~\ref{fig6}. 
In contrast, the NFC and FFC schemes do not optimize the EEI interactions, and keep on exchanging messages between robot 4 and the edge. 
This demonstrates the significance of \textbf{leveraging multi-functional designs to avoid non-beneficial collaboration for saving resources}. Note that the FFC scheme results in higher power consumption compared to the NFC scheme. 

The robots 1--3 get stuck in front of obstacles at $t=4.0\,$s, $3.9\,$s, and $8.5\,$s, respectively, 
By engaging the edge LM via NFC, these robots successfully plan executable trajectories to bypass the obstacles. 
In particular, robot 1 or 2 engages edge collaboration once, while robot 3 engages edge collaboration twice, as observed from their trajectories. 
This exactly matches the power profiles of our proposed OCN for robots 1--3. 
Note that there is a dynamic obstacle marked as blue box, which moves towards left at about $t=15$\,s. 
Before that, the corridor is totally blocked, and our method disables the edge collaboration of robot 2 until $t=18$\,s, as shown at the upper left of Fig.~\ref{fig6}. 
This demonstrates the \textbf{situation understanding} ability of OCN. 
In contrast, NFC and FFC fail to interpret the scenarios.

\section{Conclusion and Future Directions}

This article has studied the NEEI paradigm for overcoming the conflict between real-time requirements of embodied AI and high computation demands of LMs. 
The EEI-assisted NFC, NFC-assisted EEI, and resource-efficient NEEI frameworks were presented.
The REP, VBF, and OCN methods were proposed to solve the intertwined optimization between NFC and EEI functionalities. 
The frameworks and methods were verified in Carla simulation and real-world datasets.
Future directions are listed below.

\textbf{Embodied Edge Learning}. 
Robot working scenarios are highly diverse, making it impossible to enumerate all the possible cases during the model training stage. There always exists corner-case data outside the distribution of training datasets \cite{wang2023federated}. Hence, how to update the models whenever a corner case is detected, is crucial for NEEI.

\textbf{SLAM-Assisted Channel Estimation}. 
Embodied robots in NEEI can estimate their localization with ultra-high accuracy (i.e., cm-level) using SLAM. 
Thus, the angle-distance information of each near-field path can be extracted, and the NFC channel could be estimated with ultra-low pilot overhead. 

\textbf{NFC-Assisted Embodied Sensing}. 
Localization is the foundation of embodied robots. The spherical wavefront at the NFC antenna array aperture can be leveraged to estimate precisely both the angle and range of target by processing echo signals or measurements of electric field intensity. 

\bibliographystyle{IEEEtran}

\section*{Biography}

\textbf{Guoliang Li} (li.guoliang@connect.um.edu.mo) received the B.Eng. and M.Phil. (summa cum laude) degrees from the Southern University of Science and Technology, in 2020 and 2023, respectively. He is currently working toward the Ph.D. degree in computer science with the University of Macau. His research interests include autonomous systems and embodied AI.

\vspace{0.15in}

\textbf{Xibin Jin} (eexbj@mail.scut.edu.cn) received the master's degree in electronic and information engineering from the South China University of Technology. He is currently pursuing the Ph.D. degree at the South China University of Technology. His research interest includes integrated sensing and communication, motion planning, and optimization.

\vspace{0.15in}

\textbf{Yujie Wan} (yj.wan2@siat.ac.cn) received the bachelor's degree in software engineering from the Chongqing University. He is currently a Master of Computer Science and Technology at the Southern University of Science and Technology. His research interest includes robotics and radio mapping.

\vspace{0.15in}

\textbf{Chenxuan Liu} (cx.liu4@siat.ac.cn) received the B.S. degree in Electronic Information Science and Technology from Central South University, Changsha, China in 2023. He is currently working toward the M.S. degree in Computer Science and Technology with the Research Center for Cloud Computing, Shenzhen Institutes of Advanced Technology, Shenzhen, China. His research interests include robotics and 3D reconstruction.

\vspace{0.15in}

\textbf{Tong Zhang} (tongzhang@hit.edu.cn) received the B.S. degree from Northwest University, Xi’an, China, in 2012, the M.S. degree from the Beijing University of Posts and Telecommunications, Beijing, China, in 2015, and the Ph.D. degree in electronic engineering from The Chinese University of Hong Kong, Hong Kong, in 2020. He was a Postdoctoral Fellow with the Southern University of Science and Technology from 2020 to 2022. He was a Lecturer with the Department of Electronic Engineering, Jinan University. He is currently an Assistant Professor with the Harbin Institute of Technology, Shenzhen.

\vspace{0.15in}

\textbf{Shuai Wang} (s.wang@siat.ac.cn) received the Ph.D. degree from the University of Hong Kong in 2018. He is currently an Associate Professor at the Shenzhen Institute of Advanced Technology, Chinese Academy of Sciences, where he leads the Intelligent Networked Vehicle Systems (INVS) Laboratory. His research interests include autonomous systems and wireless communications.

\vspace{0.15in}

\textbf{Chengzhong Xu} (czxu@um.edu.mo) is currently a Chair Professor of Computer Science in University of Macau. His recent research focuses on cloud and edge for AI, autonomous driving and intelligent transportation.  Dr. Xu has authored two research monographs and over 600 journal and conference papers, garnering more than 22000 citations and an H-index of 79. Notably, his work has been cited in 370 international patents, including 240 U.S. patents. He is a co-inventor of more than 200 PCT and China patents and a co-founder of Shenzhen Institute of Baidou Applied Technology. He is an IEEE Fellow due to contributions in resource management in parallel and distributed computing. 

\end{document}